\documentclass[10pt,twocolumn,letterpaper]{article}

\usepackage{btas}
\usepackage{times}
\usepackage{graphicx}
\usepackage{multicol}
\usepackage{float}
\usepackage{multirow}
\usepackage{amsmath}
\usepackage{amssymb}
\usepackage{url}

\btasfinalcopy 

\ifbtasfinal\fi

\pdfoutput=1

\begin{document}


\title{On Matching Faces with Alterations due to Plastic Surgery and Disguise}

\author{Saksham Suri$^{1}$, Anush Sankaran$^{2}$, Mayank Vatsa$^{1}$, Richa Singh$^{1}$\\
$^{1}$IIIT - Delhi, India $^{2}$IBM Research, Bengaluru, India\\
{\tt\small $^{1}${\{saksham15082, mayank, rsingh\}@iiitd.ac.in}, $^{2}$anussank@in.ibm.com}
}

\maketitle

\begin{abstract}
Plastic surgery and disguise variations are two of the most challenging co-variates of face recognition. The state-of-art deep learning models are not sufficiently successful due to the availability of limited training samples. In this paper, a novel framework is proposed which transfers fundamental visual features learnt from a generic image dataset to supplement a supervised face recognition model. The proposed algorithm combines off-the-shelf supervised classifier and a generic, task independent network which encodes information related to basic visual cues such as color, shape, and texture. Experiments are performed on IIITD plastic surgery face dataset and Disguised Faces in the Wild (DFW) dataset. Results showcase that the proposed algorithm achieves state of the art results on both the datasets. Specifically on the DFW database, the proposed algorithm yields over $87\%$ verification accuracy at $1\%$ false accept rate which is $53.8\%$ better than baseline results computed using VGGFace.
\end{abstract}

\section{Introduction}
\begin{figure}[t]
	\begin{center}
		\includegraphics[width=3.3in]{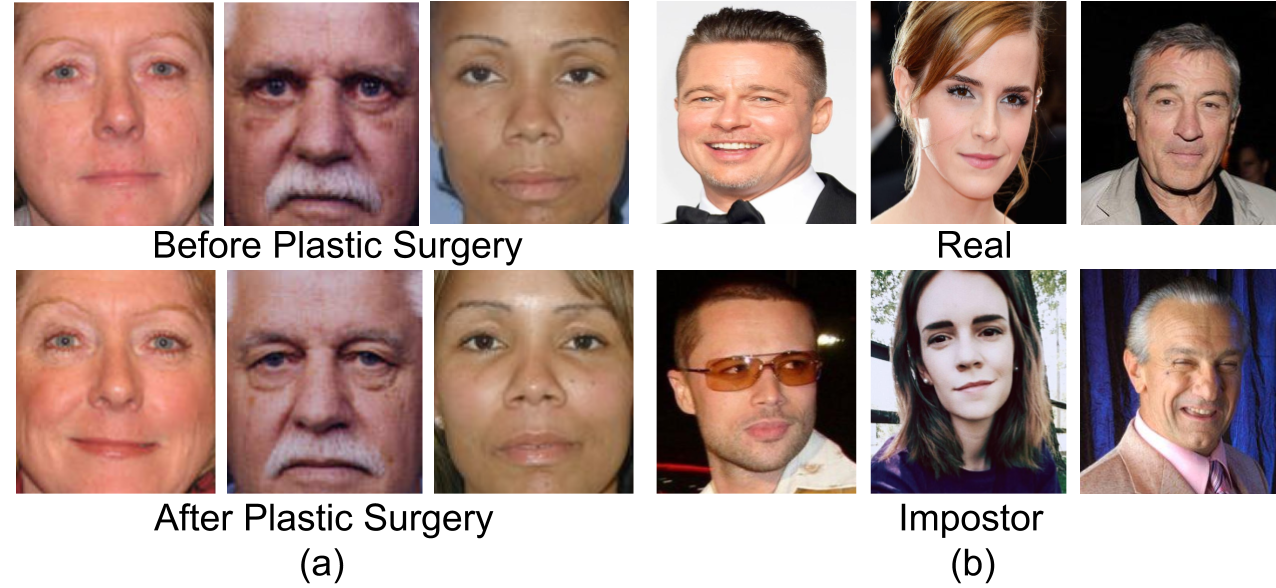}
	\end{center}
	\caption{Sample images showing two different challenges involved in face recognition: (a) plastic surgery~\cite{singh2010plastic} and (b) disguise variation~\cite{tejas2014}~\cite{dfw2018}.}
	\label{fig:motiv}
\end{figure}

Automated face recognition has been one of the breakthrough technologies of the last decade. With the advent of projects such as India's Aadhar~\cite{aadhar}, world's largest biometrics application and Apple's Face ID~\footnote{https://support.apple.com/en-in/HT208108}, face recognition technology is penetrating our day-to-day lives at a much faster rate. Challenges in face recognition are introduced by factors such as pose, illumination, expression, resolution changes, heterogeneous capture, plastic surgery, and disguise. Among these variations, plastic surgery and disguise are two of the most challenging co-variates of face recognition~\cite{bhatt2015covariates}. For example, as shown in Figure~\ref{fig:motiv} it is possible to intentionally fool a face recognition system to mask someone's original identity by wearing some disguises (temporary identity change) or undergoing a plastic surgery (permanent identity change). Hence, it is imperative and challenging to enable face recognition algorithms to cater these variations.

In literature of plastic surgery variations, Singh et al.~\cite{singh2010plastic} presented the first and only publicly available dataset, IIITD plastic surgery face dataset. Bhatt et al.~\cite{bhatt2013recognizing} proposed an algorithm for multilevel non-disjoint face granules assimilation using a multi-objective genetic approach to optimize feature extractor from each granule with weighted $\chi ^ 2$ matching. Jillela and Ross ~\cite{jillela2012mitigating} proposed a combination of information from face and ocular regions at score level. Moeini et al.~\cite{moeini2017open} developed 3D face reconstruction with sparse and collaborative representations. Most recently, Gupta et al.~\cite{gupta2018} proposed a Scattering Transform for matching surgically altered face images. There has been limited research in the field of face recognition in the presence of disguises~\cite{ dhamecha2013disguise, tejas2014, ramanathan2004facial, righi2012recognizing, singh2009face}. Recently, as part of CVPR 2018 workshop and competition, the largest publicly available Disguised Faces in the Wild (DFW) database~\cite{tejas2014}~\cite{dfw2018} was released, which contains variations due to impersonation and obfuscation. On this database, the VGG-Face model~\cite{parkhi2015deep} achieves the baseline verification results of around $33\%$ at $1\%$ False Accept Rate (FAR).
One of the major challenges in face recognition with plastic surgery and disguise is the availability of limited data. 
\begin{figure*}[!t]
	\begin{center}
		\includegraphics[width=6.2in, height = 3.3in]{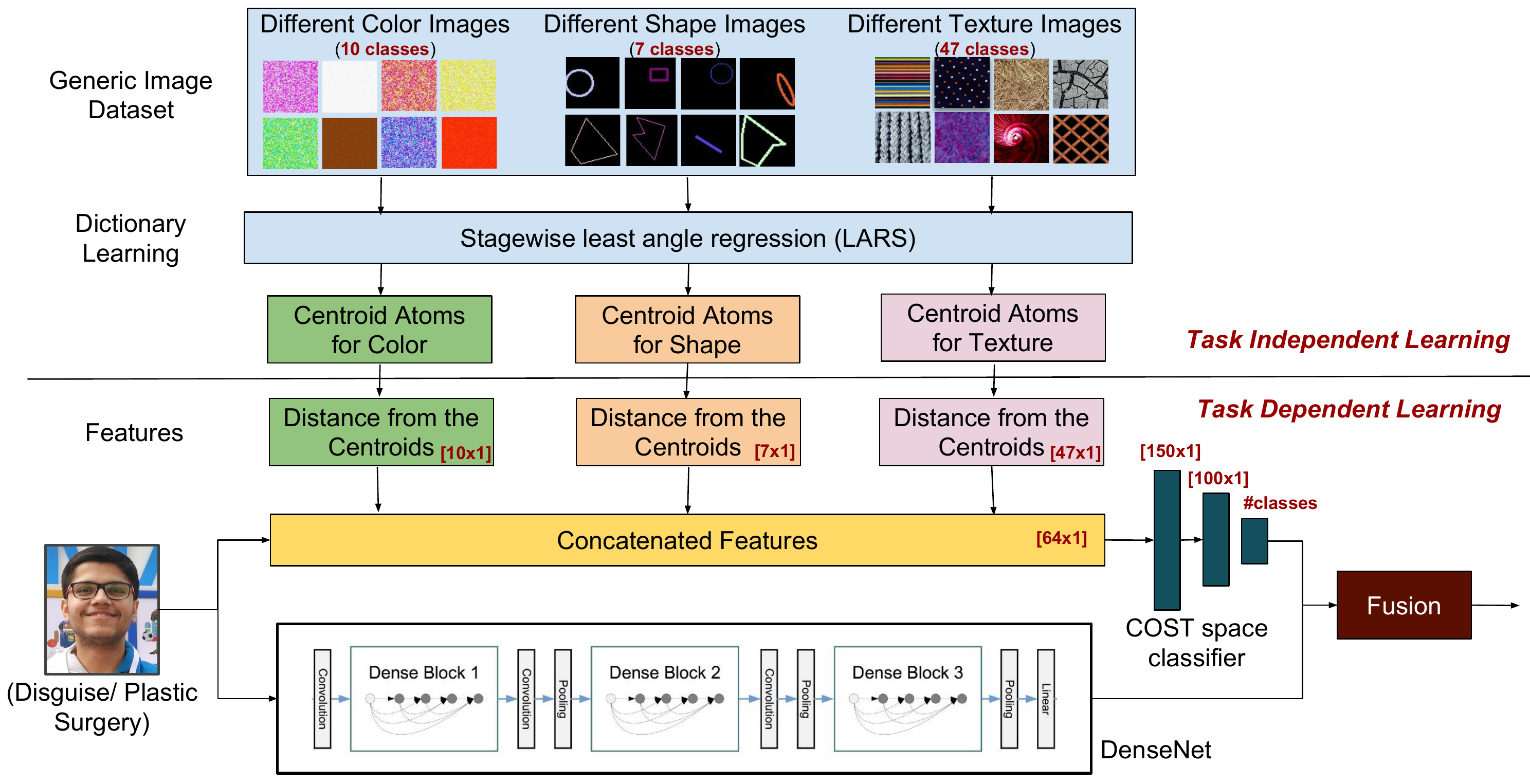}
	\end{center}
	\caption{The proposed approach of training the COST (Color (CO), Shape (S) and Texture (T)) features based classifier to supplement a task dependent supervised classifier.}
	\label{fig:approach}
\end{figure*}

To address the challenge of limited training samples, researchers have attempted transfer learning based solutions from diverse perspectives. Sankaranarayanan et al.~\cite{sankaranarayanan2017generate} performed data augmentation by simply repeating the data with small variations. Handa et al.~\cite{handa2016understanding} increased the volume of labeled data by synthetically introducing data veracity. Liu et al.~\cite{liu2015deep} bootstrapped the training by initializing the model with weights pre-trained on a similar dataset. Saenko et al.~\cite{saenko2010adapting} tried domain adaptation technique to address the lack of high volume labeled data in target domain. However, all these techniques do not consider the use of fundamental visual features, which are task independent, to boost the performance of any supervised classifier.

This paper presents a novel framework for face recognition with variations in disguise and plastic surgery. Visually, we observe that the common changes that occur in face images before and after plastic surgery or disguise is either in color, shape, or texture of the images. Thus, we proposed a novel COST  (Color (CO), Shape (S), and Texture (T)) dictionary features learnt from a generic image dataset. A classifier is trained using the COST features with the task specific labeled data. As shown in Figure~\ref{fig:approach}, the proposed framework transfers fundamental visual features learnt from a generic image dataset to supplement task specific, supervised classifiers. Experiments are performed on the benchmark datasets~\cite{tejas2014}~\cite{dfw2018}~\cite{singh2010plastic} and state of the art results are obtained using the proposed algorithm.
The rest of the paper is organized as follows: Section 2 explains the proposed framework of transferring the learnt COST features to a supervised classifier, Section 3 introduces the multiple face recognition challenges along with the dataset details. Section 4 provides the experimental results followed by conclusion in Section 5.

\section{Proposed Algorithm}
The basic principle of the proposed approach is to independently learn the representation of colors, shapes, and textures from a generic dataset. The representation is learnt using an unsupervised dictionary learning method based on stagewise least angle regression (st-LARS)  ~\cite{efron2004least} approach. Under the scenario where there is limited labeled data for a supervised classification task, two independent classifiers are trained (i) using the task specific features/models, such as the pre-trained DenseNet for face recognition, and (ii) a neural network classifier trained using the features projected on the dictionary space.

\subsection{Building the COST Space}
Let ${X_{C}}$, ${X_{S}}$, and ${X_{T}}$ be the generic image dataset on color, shape, and texture subtypes, respectively and ${Y_{C}}$, ${Y_{S}}$, and ${Y_{T}}$ be the corresponding number of classes. The purpose is to create a constrained image dataset with variations only within its subtype, such that, an unsupervised model could learn the variational representation of the subtype.

\begin{figure}[t]
	\begin{center}
		\includegraphics[width=2.76in,height=2.76in]{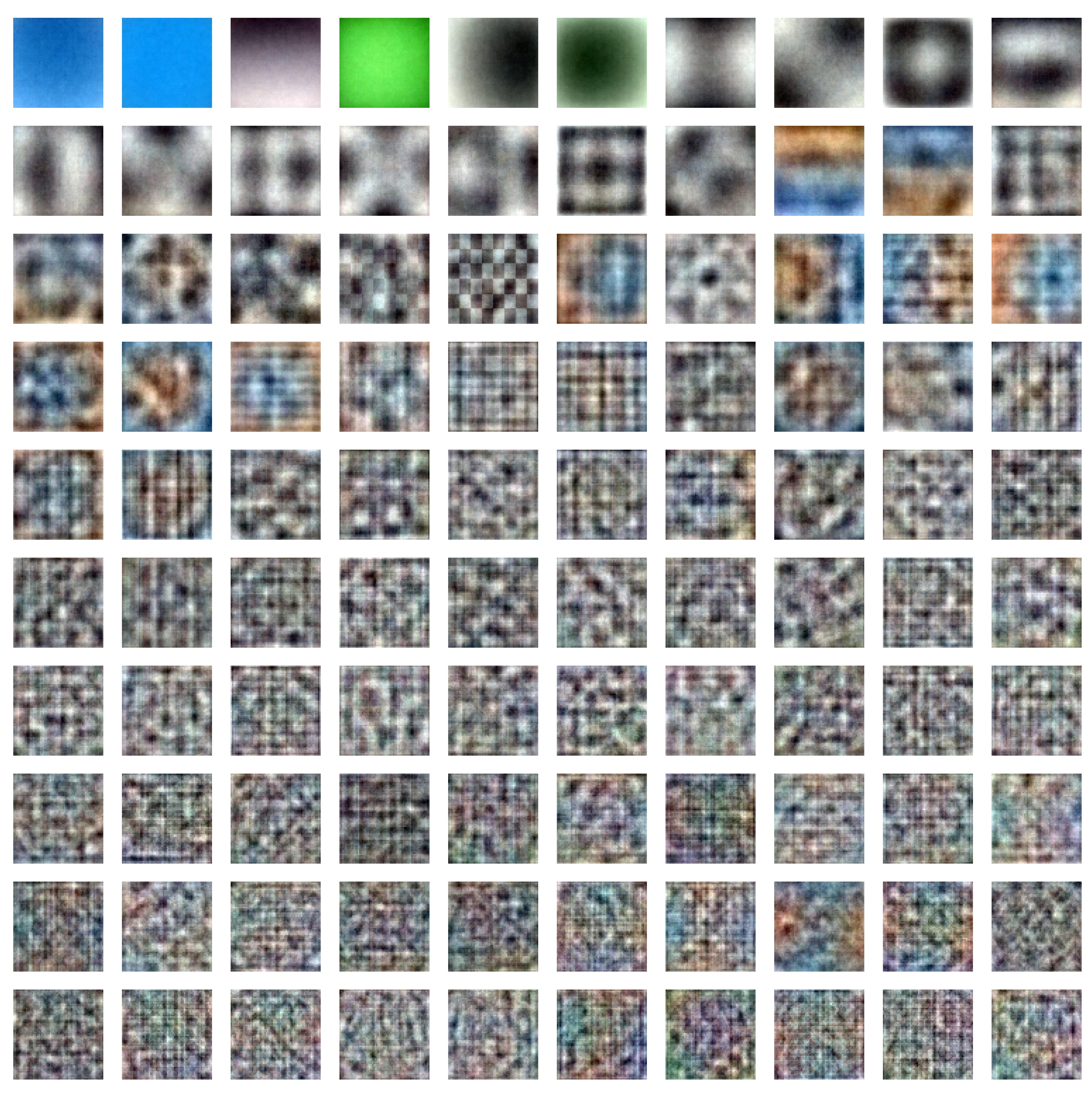}
	\end{center}
	\caption{Dictionary atoms learnt from the texture images subset.}
	\label{fig:dict}
\end{figure}

\subsubsection{Color dataset}
For the color subtype, we used $10$ classes (${Y_{C}}$) namely: red, green, blue, yellow, magenta, cyan, black, white, brown, and orange. The images are generated pixel-wise such that each pixel will have a (R, G, B) within a constrained range of the base class color. For example, while generating an image of class ``red'', every pixel is chosen as a random (R, G, B) value in the range ($200-255$, $0-255$, $0-255$). Thus, predominantly the image would have a red color with speckle noise to introduce variations while learning the representation of red color. Each image is of size ($250\times250\times3$) and $1,000$ images are generated per class, creating a total of $10,000$ images for the color subtype.
\subsubsection{Shape dataset}
For the shape subtype, we used $7$ classes (${Y_{S}}$) namely: lines, rectangle, circle, ellipse, quadrilateral, pentagons, hexagons. On a black image, the shapes are generated with varying color boundaries (10 colors), varying locations on the image, varying perimeter, varying angle if possible and also varying boundary thickness ($1-5$ pixels). Each image is of size ($250\times250\times3$) and $1,000$ images are generated per class, creating a total of $7,000$ images for the shape subtype.
\subsubsection{Texture dataset}
For the texture subtype, we used the Describable Texture Dataset (DTD) ~\cite{cimpoi2014describing} which contains $5,640$ images from $47$ different textures (${Y_{T}}$) with $120$ images per class. The images vary in size from ($300\times300\times3$) to ($640\times640\times3$), with at least $90\%$ of the image describing the corresponding texture. All the images are re-sized to ($250\times250\times3$) for our experiments.

These three datasets are utilized to learn the basic representation of colors, shapes, and texture in the visual domain. While this research work focuses on these three subtypes, an obvious extension is to include additional subtypes and additional classes within each subtype.

\subsection{Learning the COST Dictionary}

For an object classification task, we aim to extract three basic visual cues - color, shape, and texture. The aim is to learn a COST feature space representation using different colors, shapes, and texture through an unsupervised learning method. A supervised classifier could be independently trained over these COST features, which learns the mapping of color $c_i$, shape $s_i$, and texture $t_i$ to the object class. This supervised classifier could be trained for any task and for any dataset due to the generic, task independent nature of the features. For learning the dictionary feature from the color images, the optimization function is described as follows,
\begin{equation}
min \left[\frac{1}{M} \sum_{i=1}^{M}  min_{h_C^{(i)}} \underbrace{||X_C^{(i)} - \overbrace{D_C}^\text{Dictionary} h_C^{(i)}||_2^2}_\text{Reconstruction Error} +  \lambda_C \underbrace{||h_C^{(i)}||_1}_\text{Sparsity} \right]
\end{equation}

where, $X_C^{(i)}$ represents the $i^{th}$ image sample,  $D_C$ is the learnt dictionary, $h_C^{(i)}$ is the feature representation learnt for the $i^{th}$ image sample, $\lambda_C$ is the sparsity controlling parameter, $||.||_p$ denotes the $\ell_p$-norm, and $M$ is the total number of samples used for training. The function learns the sparse representation $h_C^{(i)}$ and the dictionary model $D_C$ that minimizes the overall reconstruction error. 

In the dictionary learning approach, while $\ell_0$-norm could achieve an ideal sparsity solution, it is not differentiable and hence the optimization function becomes NP-hard. The basis pursuit ~\cite{chen2001atomic} and LASSO ~\cite{tibshirani1996regression} are two popular greedy approaches used to replace the $\ell_0$-norm with $\ell_1$-norm, but with the trade off of having a high computational complexity. Thus, we adopt the idea of st-LARS (Stagewise Linear Angle Regression) to approximate using a greedy technique but in linear computational time. A similar optimization function is used to learn the dictionary representation of shape and texture images. Figure~\ref{fig:dict} shows a visualization of the dictionary learnt for the texture dataset.

Note that any unsupervised feature learning approach could provide a similar optimization function, as provided in Equation (1), and thus can be interchangeably used. The primary advantage of using a dictionary learning based approach is that it encodes $h_C^{(i)}$ as a complex function of the input $X_C^{(i)}$ as follows~\cite{mairal2009online}:
\begin{equation}
h_C(X_C^{(i)}) = argmin_{h_C^{(i)}} ||X_C^{(i)} - D_C h_C^{(i)}||_2^2 + \lambda_C ||h_C^{(i)}||_1
\end{equation}

Using the training images, features are obtained independently for color, shape, and texture images. From these color, shape and texture images the centroids of each class (10 in case of color) are computed. A total of $64$ centroids, i.e. $10$ for color, $7$ for shape, and $47$ for texture are obtained. Any image can now be represented as a fixed length vector $64\times1$ as its Euclidean distance from these $64$ centroids after getting the coefficients of the image corresponding to the learnt dictionaries. Using the limited labeled data and shape, color, and texture feature models, we train a two hidden layer, neural network based classifier. As the features are the distance from the class centroids, they represent the dominant colors, shapes, and texture in the image which the neural network utilizes for the classification task at hand. Thus, these features could be used to supplement any task specific classifier that are learnt on top of a human engineered or automatically learnt features.

\subsection{DenseNet: Task Dependent Supervised Classifier}

To show the generic nature of the proposed framework, we choose an off-the-shelf deep learning model, DenseNet ~\cite{huang2017densely}, as the task dependent supervised classifier. The DenseNet is pre-trained on the ImageNet dataset ~\cite{deng2009imagenet} and is further fine-tuned on the different datasets used in this research. DenseNet is one of the state-of-art deep learning models for object classification and thus, an improvement to this model by the proposed framework can showcase the effectiveness of the COST based learning. In this research, the DenseNet-121 having 121 trainable layers with three dense blocks, is useful in extracting highly local and complex features from the given input image. As shown in Figure~\ref{fig:approach}, each dense block in DenseNet consists of a sequence of convolution layer, where every layer takes as input all the preceding layers' response within that block. 

Two kinds of classification experiments are performed to show the diversity of proposed framework: (i) identification is an \textit{n-class} classification setting, where the input image is classified to one of the available classes, and (ii) verification is a binary classification setting, where two images are compared to verify if they belong to the same class or not. The last fully connected layer of the DenseNet is removed and replaced with a fully connected layer with number of nodes equal to the number of classes in case of identification or two nodes in case of verification.

\subsection{Classifier Fusion}
In this section, we present the classifier fusion approaches for verification and identification scenarios.

\textbf{Verification: }
To perform verification, distance between the two images of a pair is calculated using the softmax activations of the corresponding network (separately for COST dictionary based neural network and DenseNet). A weighted sum of the distances computed using COST dictionary based neural network's output and DenseNet based features is computed. The equation of score fusion is written as follows:
\begin{equation}
\label{eq:eqn13}
{dist}_{new}^{(i)} = \alpha.{dist}_{cost}^{(i)} + (1-\alpha).{dist}_{supervised}^{(i)}
\end{equation}
where ${dist}_{cost}^{(i)}$ is the COST feature space based distance, ${dist}_{supervised}^{(i)}$ is the distance calculated using the output of the supervised classifier and ${dist}_{new}^{(i)}$ is the combined distance. $\alpha$ is used to decide the weights given to the two distances being combined.

\textbf{Identification: }
For performing identification, score of a probe is calculated with respect to all images present in the gallery set. Based on the distances obtained, the rank at which each sample is correctly identified is computed. These identification accuracies are then used for computing the Cumulative Match Characteristic (CMC) curve.

\subsection{Implementation details}
Color, shape and texture dataset images are resized to $(64\times 64\times 3)$ for dictionary learning based feature extraction and face images are resized to $(224\times 224\times 3)$ for DenseNet based feature extraction. The dictionary learning algorithm is executed for $100$ epochs, the COST feature based neural network classifier is executed for $20,000$ epochs, and DenseNet is finetuned for $100$ epochs. For classifier fusion, the $\alpha$ parameter is obtained through extensive grid search as $\alpha$ = $0.3$.

\section{Datasets and Protocols}
In this research, we show the results of the proposed framework on two different datasets:
\begin{enumerate}
    \item \textbf{Plastic Surgery Face Dataset: } This is a real world dataset with $1,800$ pre- and post-surgery images corresponding to $900$ subjects. The alterations present in the dataset~\cite{tejas2014}~\cite{dfw2018} include browlift, facelift, skin tone change, nose-job, liposuction, ear alterations, fat injection, lip alterations, eyelid alterations, and chin modifications.  The experiments are performed using the original protocol~\cite{singh2010plastic} with 10 times cross-validation with $40\%$ data in training set and $60\%$ data in the testing set for each fold.
    
    \item \textbf{Disguised Faces in the Wild (DFW) Dataset: }
    The dataset consists of $1,000$ subjects and total of $11,155$ images. As per the pre-defined protocols, $400$ subjects comprise the training set and $600$ subjects comprise the testing set. The dataset has four types of images namely normal, validation, disguised and impersonator face images. Face coordinates generated from Faster-RCNN have been provided along with the images. Three protocols for reporting the results have been provided with the dataset. Protocol-1 can be used for evaluating an algorithm under impersonation only and uses pairs formed using normal and validation images of the same subject as genuine pairs and those made using normal, validation and impersonator images of the same subject as imposter pairs. Protocol-2 can be used for evaluating an algorithm for disguises via obfuscation only and uses pairs having at least one disguise image and the other as normal, validation or disguise image of the same subject as genuine pairs while cross subject pairs generated using the normal, validation and disguised images of another subject as the imposter set. Protocol-3 uses the entire dataset for evaluating the algorithm. Genuine pairs are formed by combining the pairs created in the above two protocols. The imposter pairs are created using the impersonator images with the normal, validation, and disguised images of the same subject, along with cross-subject imposter pairs.
\end{enumerate}

\begin{figure}[!t]
	\begin{center}
		\includegraphics[width=3.6in,height=2.5in]{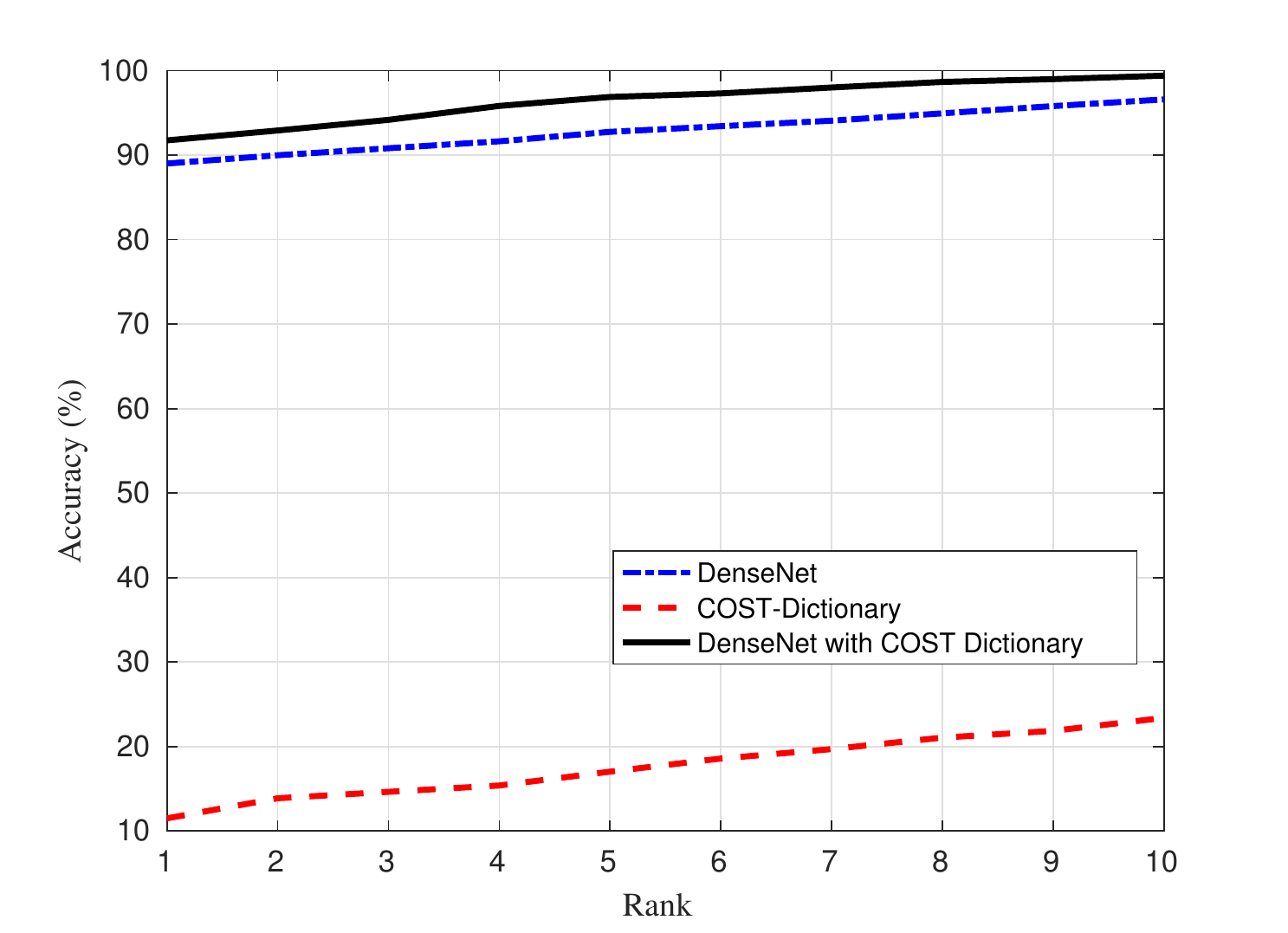}
	\end{center}
	\caption{CMC curve for the proposed algorithm on the IIITD Plastic Surgery Face dataset}
	\label{fig:cmc}
\end{figure}

\begin{figure}[!t]
	\begin{center}
		\includegraphics[width=3.6in,height=2.5in]{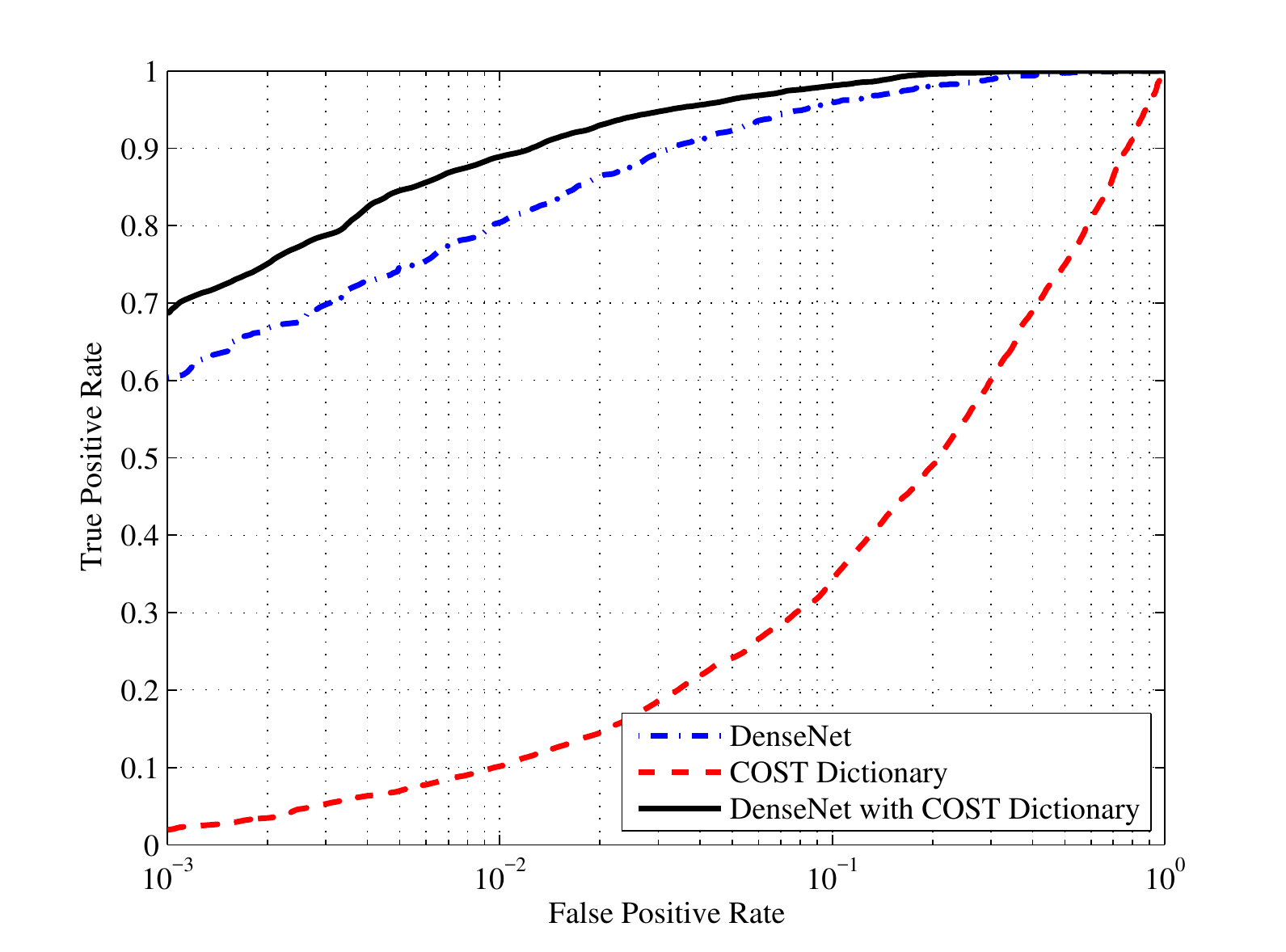}
	\end{center}
	\caption{Face verification using the proposed algorithm on the IIITD Plastic Surgery Face dataset}
	\label{fig:roc_ps}
\end{figure}

\begin{table}[t]
\centering
\begin{tabular}{|p{35mm}|l|l|l|}
\hline
Algorithm   & Rank1 & Rank5 & Rank10 \\ \hline
\hline
TPLBP~\cite{wolf2008descriptor}* & 70.33  & 85.33  & 88.70    \\ \hline
Bhatt et al.~\cite{bhatt2013recognizing}*  & 87.32  & 92.05  & 97.26   \\ \hline
I. Gupta ~\cite{gupta2018}* & 85.43  & 95.91  & 97.61   \\ \hline
COST Dictionary                          & 11.49  & 17.02  & 23.39   \\ \hline

DenseNet                                & 89.01  & 92.76  & 96.60    \\ \hline
DenseNet + COST Dictionary (Proposed)                                & \textbf{91.75}  & \textbf{96.89}  & \textbf{99.41}   \\ \hline
\end{tabular}
\caption{Results on IIITD Plastic Surgery Dataset. Results marked with * were taken from the corresponding papers.}
\label{ps_res}
\end{table}

\section{Results and Analysis}
The effectiveness of the proposed framework is evaluated on IIITD Plastic Surgery Face dataset and DFW database. The COST space representation for each image is calculated by finding the distance between the centroids and the extracted features. For the combination of results from DenseNet and COST space based representation, the softmax activations from DenseNet and dictionary are combined in the manner stated in section 2.4. The results for the two datasets are analyzed in the sections below.

\subsection{IIITD Plastic Surgery Face dataset}
Figure ~\ref{fig:cmc} shows the CMC curve for the proposed algorithm on the IIITD Plastic Surgery Face dataset along with separate plots for COST based and DenseNet based scores. Table ~\ref{ps_res} gives a comparative study of the proposed algorithm along with state of the art algorithms on the IIITD Plastic Surgery Face dataset. We observe around 4.5\% improvement in the Rank-1 accuracy over the current best reported results on the dataset. Although COST based dictionary alone does not perform well, but combined with DenseNet, an improvement of around 2\% is observed in the Rank-1 accuracy. Figure ~\ref{fig:roc_ps} shows the ROC curve for verification performance of the proposed algorithm. The primary reason for this improved performance is due to better learning the primitive shape, color and texture features. 
\begin{figure}[!t]
	\begin{center}
		\includegraphics[width=3.6in]{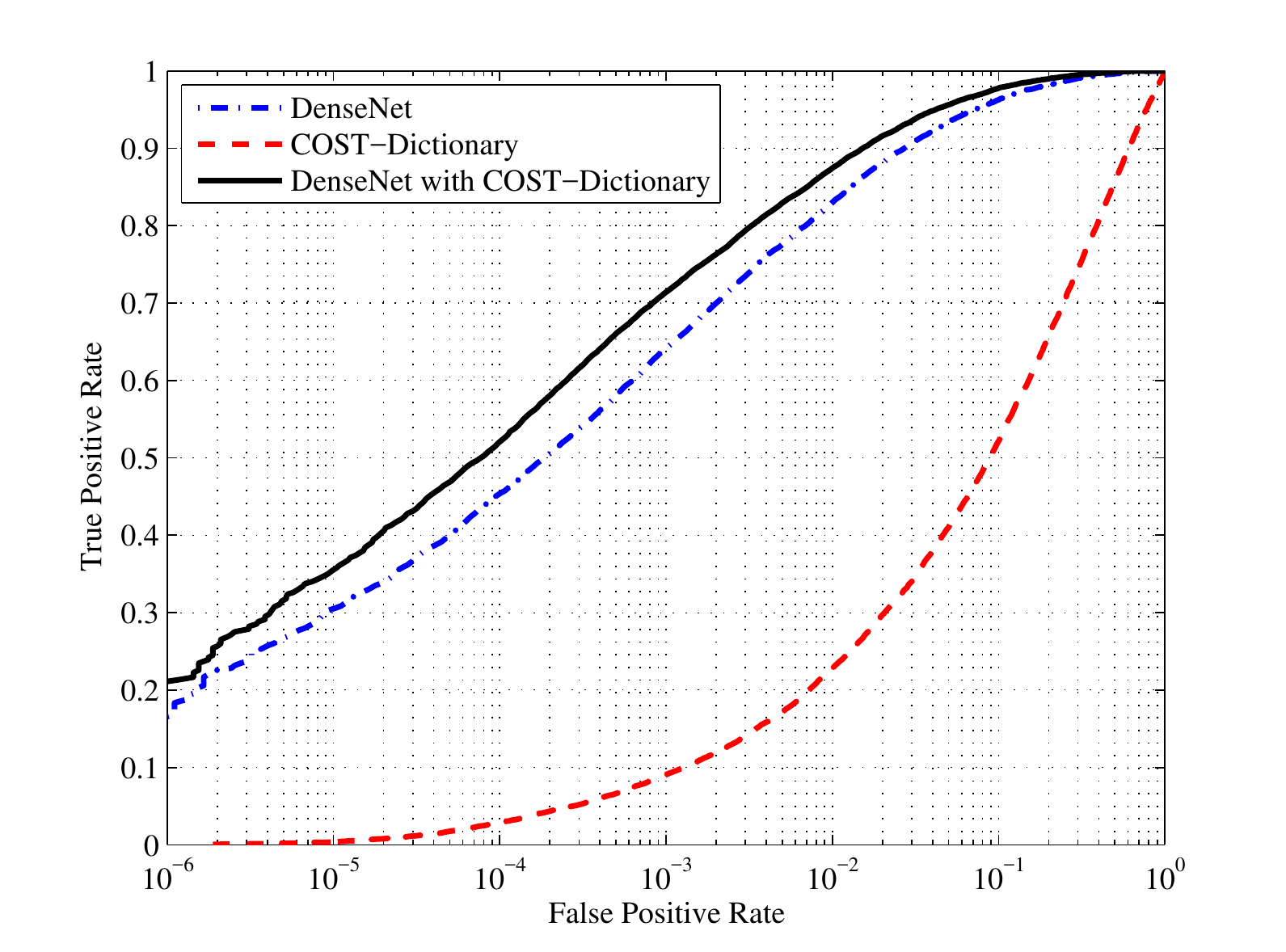}
	\end{center}
	\caption{Face verification using the proposed algorithm on the DFW dataset}
	\label{fig:roc_dfw}
\end{figure}

\begin{table*}[!t]
	\begin{center}
		\begin{small}
			\begin{tabular}{|l|c|c|c|c|c|c|}
				\hline
				\multirow{2}{*}{Algorithm} & \multicolumn{3}{|c|}{1\% FAR} &\multicolumn{3}{|c|}{0.1\% FAR} \\ \cline{2-7}
				 & Protocol 1 & Protocol 2 & Protocol 3 & Protocol 1 & Protocol 2 & Protocol 3\\
				\hline \hline
				VGG-Face (Baseline)* & 52.8 & 31.5  & 33.8 & 27.1 & 15.7 & 17.7 \\ 
				\hline

				COST Dictionary  & 27.3  & 22.4 & 22.9 & 12.7 & 8.5  & 9.0   \\ \hline
			
				DenseNet &	89.8  & 82.9 & 83.1  &  59.6 & 64.4   & 64.1   \\ \hline
				DenseNet + COST Dictionary (Proposed) &	\textbf{92.1} & \textbf{87.1} & \textbf{87.6}  & \textbf{62.2} & \textbf{72.1} & \textbf{71.5}   \\ \hline
			\end{tabular}
		\end{small}
		\caption{\label{dfw_res} Verification accuracy at 1\% and 0.1\% FAR on the DFW dataset. Results for algorithm marked with * were provided with the dataset and have not been computed by the authors.}
	\end{center}
\end{table*}

\subsection{DFW dataset}
For DFW dataset, the results have been compared at 1\% FAR and 0.1\% FAR as per the protocol for three different scenarios. The Table ~\ref{dfw_res} summarizes the results. Figure ~\ref{fig:roc_dfw} shows the ROC for Protocol-3 which uses the entire dataset for evaluation. The proposed algorithm outperforms the baseline approach for all protocols. For Protocol-3 which evaluates the algorithm on the entire dataset, an improvement of around 53.8\% and 53.8\% is observed from the baseline at 1\% and 0.1\% FAR, respectively. Also combining DenseNet and COST based dictionary predictions improves the overall performance by 4.5\% and 7.4\% at 1\% and 0.1\% FAR respectively as compared to only DenseNet based predictions. Further, comparing the results from the DFW competition~\cite{dfw2018}, the proposed algorithm yields the second best results.

\begin{figure}[!t]
	\begin{center}
		\includegraphics[width=3in, height=2in]{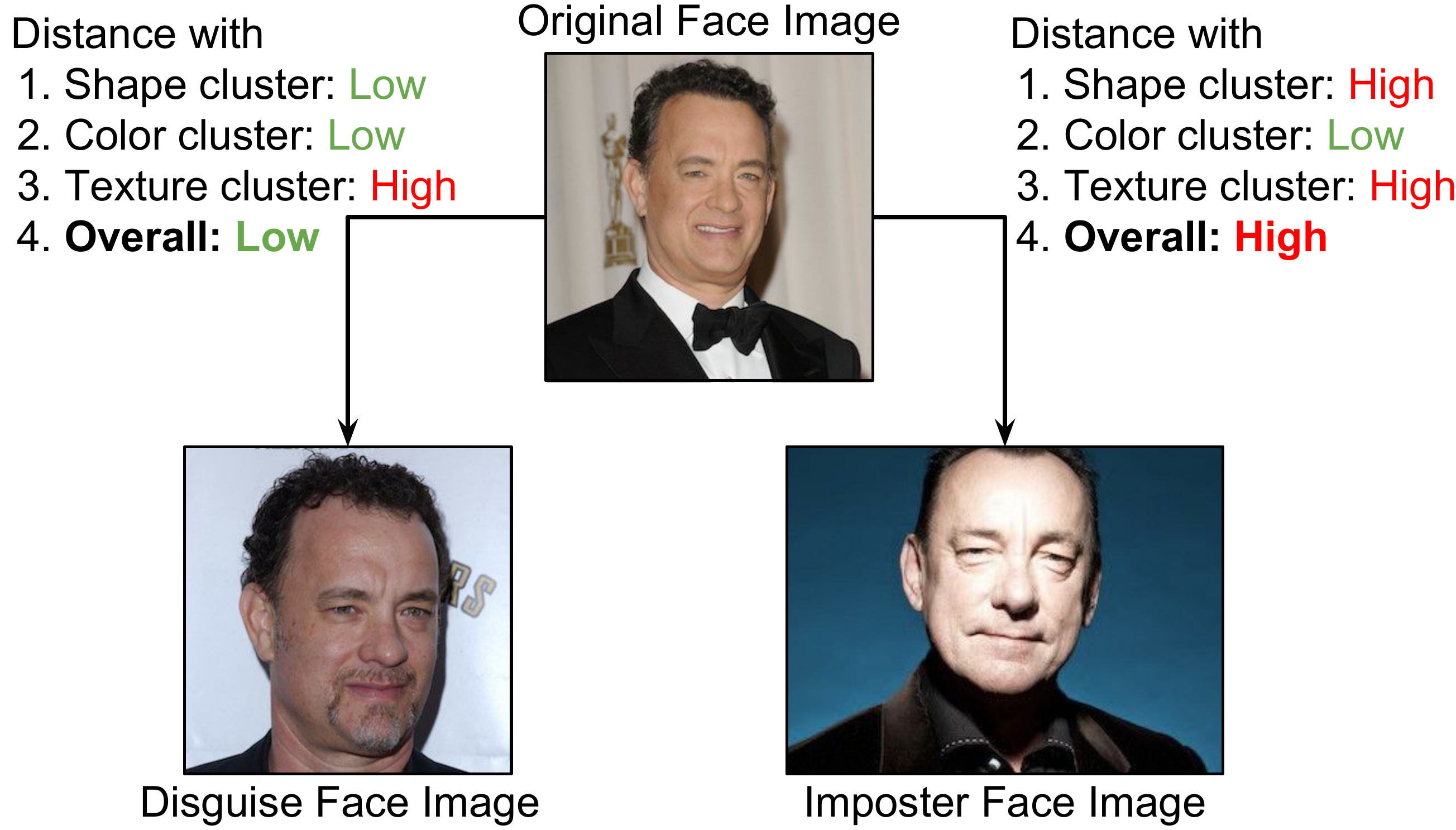}
	\end{center}
	\caption{Example to showcase the necessity of adding COST based representation to a supervised classifier.}
	\label{fig:obs}
\end{figure}

\subsection{Observations}
COST feature space in the proposed framework captures the high level
meta information of the given face image that aids in the classification task. A deep learning model such as DenseNet
would capture the complex relationship between the pixels
in its hidden layers, while supplementing basic visual cues
and meta information, could enrich the feature representation.
Consider the celebrity face of \textit{Tom Hanks}, as shown in Figure~\ref{fig:obs}, as an example to describe the value addition of the COST features. Visually, the imposter face image and the disguise face image of \textit{Tom Hanks} might not look very different from the original image. However, the distance of the original image from the genuine and imposter image in the color, shape, and texture clusters provides more interpretable information. The disguise face image has a higher distance only in the texture cluster suggesting that there is much variation only in the texture between the two images. While the imposter image has high variation in shape, texture, and overall distance. Thus, the idea of COST feature space is to capture the meta information from a face image and supplement it with the complex deep learning model, to improve the overall performance of face recognition and classification.

\section{Conclusion and Future Work}
In this research, we proposed a framework for learning and using visual cues such as shape, color and texture for image classification task. The representation for shape, color, and texture are learnt using unsupervised dictionary learning from a carefully curated generic image dataset. The usefulness of the algorithm is studied across two different face alteration datasets. Experimentally we showed that supplementing DenseNet-121 with the proposed COST space classifier improved the performance of the overall framework. As further improvement we aim to include saliency to the COST space representation. It may further improve the performance as it would help in encoding the position and localization of important parts of the images to the pipeline as well.

\section{Acknowledgements}
Vatsa and Singh are partly supported through Infosys Center for Artificial Intelligence at IIIT-Delhi.

{\small
\bibliographystyle{ieee}
\bibliography{btas_intuition}
}

\end{document}